\documentclass[10pt,twocolumn,letterpaper]{article}

\usepackage{cvpr}
\usepackage{times}
\usepackage{epsfig}
\usepackage{graphicx}
\usepackage{amsmath}
\usepackage{amssymb}
\usepackage{multirow}
\usepackage{bm}
\usepackage{array}
\usepackage{hhline}
\usepackage{subfigure}
\usepackage{caption} 
\usepackage{float}

\newcolumntype{M}[1]{>{\centering\arraybackslash}m{#1}}

\usepackage[pagebackref=true,breaklinks=true,letterpaper=true,colorlinks,bookmarks=false]{hyperref}

\cvprfinalcopy 

\def\cvprPaperID{1514} 

\ifcvprfinal\fi
\begin{document}

\title{Transferring a Semantic Representation for Person Re-Identification and Search}

\author{Zhiyuan Shi, Timothy M. Hospedales, Tao Xiang\\
Queen Mary, University of London, London E1 4NS, UK\\
{  $\{$z.shi,t.hospedales,t.xiang$\}$@qmul.ac.uk}
}
\maketitle
\begin{abstract}

Learning semantic attributes for person re-identification and  description-based person search has gained increasing interest due to attributes' great potential as a pose and view-invariant representation. However, existing attribute-centric approaches have thus far underperformed state-of-the-art conventional approaches. This is due to their non-scalable need for extensive domain (camera) specific annotation.  In this paper we present a new semantic attribute learning approach for person re-identification and search. Our model is trained on existing fashion photography datasets -- either weakly or strongly labelled. It can then be transferred and adapted to provide a powerful semantic description of surveillance person detections, without requiring any surveillance domain supervision. The resulting representation is useful for both unsupervised and supervised person re-identification, achieving state-of-the-art and near state-of-the-art performance respectively. Furthermore, as a semantic representation it allows description-based person search to be integrated within the same framework.

\end{abstract}


\vspace{-3pt}

\section{Introduction}

Person re-identification (re-id) and description-based search are crucial tasks in visual surveillance. They underpin many fundamental applications including multi-camera tracking, crowd analysis and forensic search. Both tasks aim to retrieve images of a specific person, but  differ in the query used. Person re-identification queries using an image from a different view (e.g., in multi-camera tracking), while person search uses a textual person description (e.g., eyewitness description). Despite extensive research \cite{gong2014reidBook,vezzani2013reidSurvey}, these tasks remain unsolved due to various challenges including the variability of viewpoints, illumination, pose, partial occlusion, low-resolution and motion-blur  \cite{gong2014challenge}. 

Attributes, as a mid-level semantic representation, could potentially address these shared challenges, as they are intrinsically invariant to these variabilities in viewing conditions. Attribute-centric approaches are inspired by the operating procedure of human experts, and revolve around the idea of learning detectors for person attributes (such as Blue-Jeans).  In particular, a number of approaches are developed for person re-identification using an attribute-based representation \cite{layne2013attribReIdBook,roth2014jointAttributeLearning,li2014clothesAttrib}. It was demonstrated \cite{layne2013attribReIdBook} that 
 \emph{if} the each person's attributes can be reliably detected,  re-id is essentially solved (since a $K$ binary attribute descriptor can potentially disambiguate $2^K$ people). For person search, since most textual descriptions used for query are attributes, an attribute-based approach is often the only option.   
 
Despite their hoped-for potentials, attribute-centric approaches to person re-identification have until now not achieved state-of-the-art performance compared to conventional alternatives focused on learning effective low-level features and matching models \cite{vezzani2013reidSurvey}. We argue that this is largely due to the difficulty of obtaining sufficient and sufficiently detailed annotations to learn robust and accurate attribute detectors. In particular, to achieve good attribute detection, per camera/dataset annotations need to be obtained, since attribute models typically do not generalise well across cameras (e.g.~a blue shirt may look purple in a different camera view). This is exacerbated, because unlike annotation of person identity used in learning a re-id matching model, attributes require multiple labels per image. Moreover since most human attributes are associated with a specific body part, to learn accurate attribute detectors, ideally annotation needs to be done at the patch-level rather than the image-level. In short, an attribute-based approach is limited by the scale of its annotation requirements.

Although attribute-centric approaches to surveillance are limited by the lack of annotation, extensive person attribute annotations already exist in other domains, notably fashion image analysis \cite{Liu_MM_2014,Yamaguchi2012,Yang_2014_CVPR,Yamaguchi_2013_ICCV,Liu2012}. In this line of work, many high resolution images have been annotated with clothing properties -- sometimes strongly (per-pixel). However, learning attribute detectors from the fashion domain, and applying them directly to person re-id and search will fail, due to the severe domain shift problem -- compared with   surveillance data, the  image characteristics are completely different (see Fig.~\ref{fig:what is learned}). In particular,  many of the challenges in surveillance are absent (e.g.~illumination variability, occlusion, low-resolution, motion blur).  These large and perfectly annotated person attribute datasets are thus useless, unless an attribute model learned from them can be successfully adapted and transferred to the surveillance domain.

In this paper we contribute a new framework that is capable of learning a semantic attribute model from existing fashion datasets, and adapting the resulting model to facilitate person re-identification and search in the surveillance domain. In contrast to most existing approaches to attribute detection \cite{layne2013attribReIdBook,roth2014jointAttributeLearning} which are based on discriminative modelling, we take a generative modelling approach based on the Indian Buffet Process (IBP) \cite{Griffiths_2011}. The generative formulation provides  key advantages including: joint learning of all attributes; ability to naturally exploit weakly-annotated (image-level) training data; as well as unsupervised domain adaptation through Bayesian priors. Importantly a IBP-based model \cite{feng2014learnDict,shi2014wslAttrib,Konstantina_tpami_2014} provides the favourable property of combining attributes factorially in each local patch. This means that our model can differentiate potentially ambiguous situations such as Red-Shirt+Blue-Jeans versus Red-Jeans+Blue-Shirt (See Fig.~\ref{fig:person search}). Moreover, with this representation, attribute combinations that were rare or unseen at training time can be recognised at test time so long as they are individually known (e.g.~Shiny-Yellow-Jeans).

Our framework overcomes the significant problem of domain shift between fashion and surveillance data in an unsupervised way by Bayesian adaptation. It can exploit both strongly and weakly annotated source data during training, but is always able to produce a strong (patch-level) attribute prediction during testing. The resulting representation is highly person variant while being view-invariant, making it ideal for person re-id, where we obtain state-of-the-art results. Moreover, as the representation is semantic (nameable or describable by a human), we are able to unify description based person search within the same framework, where we also achieve state-of-the-art results. 

\section{Related Work}

\noindent \textbf{Person Re-identification:}
Person re-identification is now a very well studied topic, with \cite{gong2014reidBook,vezzani2013reidSurvey} providing good reviews. 
Existing approaches can be broadly grouped according to two sets of criteria: supervision and representation/matching. Unsupervised approaches \cite{zhao2013unsupervised,Farenzena_2010CVPR,wang_2014_bmva} are more practical in that they do not require the use of per-target camera annotation (labelling people into matching pairs), while supervised approaches \cite{SalientColor_eccv_2014,abir_eccv_2014,xiong_eccv_2014,bischof2012largeScaleMetricLearn,zhao2013person,Kuo_2013}  that use this information tend to perform better. Studies also either focus on building a good representation \cite{Farenzena_2010CVPR,VIPeR_2008,zhao2013unsupervised,Pedagadi_2013_CVPR,li2014deepreid,wang_2014_bmva,SalientColor_eccv_2014}, or building a strong matching model \cite{xiong_eccv_2014,Liu_2014_CVPR,Xu_2013_ICCV,ZhenliShiyu_CVPR2013,bischof2012largeScaleMetricLearn,zhao2013person}. 
In this paper we focus on learning mid-level semantic representations for both unsupervised and unsupervised re-id. Zhao et al.~\cite{zhao2014learning}  also attempted to learn mid-level representations. However their method does not learn a \emph{semantic} representation, so it cannot be used for description-based person search; and it relies on supervised learning. Ours can be used with or without supervision, but provides the biggest benefit in the unsupervised context. A few studies \cite{Ma_2013_ICCV,cvpr_ZhengGX12} address person re-id based on transfer learning. However, they transfer matching models rather than representations, and between different surveillance datasets rather different domains.

Semantic attribute-based representations have also been studied for person re-identification. A key motivation is that it is hard for low-level representations (e.g., colour/texture histograms) to provide both  view-invariance and person-variance. 
Early studies train individual attribute detectors independently on images weakly-annotated with multi-label attributes \cite{layne2013attribReIdBook,liu2012attribTopic,Ryan_bmvc_2014}, followed by using the estimated attribute vector of a test image as its representation for matching. More recent studies have considered joint learning of attributes and inter-attribute correlation \cite{roth2014jointAttributeLearning,li2014clothesAttrib,khamis2014jointAttribute}. Nevertheless, these approaches do not reach state-of-the-art performance. This is partly due to the drawbacks of (1) requiring ample target domain attribute annotations, which are hard to obtain at scale for learning robust detectors; and (2) producing coarse/weak (image-level) rather than strong (patch-level) annotations -- due to the lack of strongly annotated training data that is even harder to obtain. In contrast, by transferring  an attribute representation learned on existing large annotated fashion datasets, we  overcome these issues and obtain state-of-the-art results. 

\noindent \textbf{Person Search:}
Related to attribute-based person re-identification is description-based person search \cite{vaquero2009attrib_surveil,Feris_icmr_2014,layne2013attribReIdBook,Ferrari09,satta2012appearanceSearch}. In this case detectors for aspects of person description (e.g., clothes, soft-biometrics) are trained. Person images are then queried by description rather than by another image as in re-id. Most methods however have limited accuracy due to: (i) training on, and producing weak (image-level) annotations; and (ii) training and testing each attribute detector independently rather than jointly. For these reasons, they are also limited in being able to make complex attribute-object queries such as ``Black-Jeans + Blue-Shirt'', which requires a strong joint   segmentation and attribute model to disambiguate  the associations between attributes.

Similar notions of attribute-based search have been exploited in a face image context \cite{kumar2008faceTracer}; and in a fashion/e-commerce context, where users search for clothing by description \cite{Yamaguchi_2013_ICCV,Huizhong2012,Liu_MM_2014,BourdevAttributesICCV11}. Face image search however, is easier due to being more constrained and well aligned than  surveillance images. Meanwhile in the fashion context, clean high resolution images are assumed; and crucially strong pixel-level annotations are typically used, which would be prohibitively costly to obtain for  individual surveillance camera views. In this paper we bridge the gap between the clean and richly annotated fashion domain, and the noisy and scarcely annotated surveillance domain, to produce a powerful semantic representation for person re-id and search.

\noindent \textbf{Domain Adaptation:}
A key challenge for our strategy is the domain-shift between the fashion and surveillance domains. Addressing a change of domains with domain adaptation is well studied in computer vision \cite{fernando2013SA,saenko2010domainAdapt,Cross_2012_wei} and beyond \cite{pan2009transfer_survey}. To avoid necessitating target domain annotation, our task requires unsupervised adaptation which is harder. Some off-the-shelf solutions exist \cite{fernando2013SA,gong2012geodesicFlowDA}, but these under-perform due to working blindly in the low-level feature space, disconnected to the semantics being modelled, i.e.~attributes. In contrast, in the style of \cite{mccallum1998activeem,cao2010crossdata_action,Cross_2012_wei}, we achieve domain adaptation by transferring the source domain attribute model as a prior when the target domain model is learned. This enables adaptation to the target domain, while exploiting the constraints provided by semantic attribute model.
  
\noindent \textbf{Attribute Learning:}
Beyond person re-id and search, semantic attribute learning \cite{NIPS2007_3217,Zitnick_2013_ICCV} is well studied in computer vision, which is too broad to review here. However, we note that most attribute-learning studies have one or more simplifications compared to the problem we consider here: They consider within-domain rather than cross-domain attributes; or produce coarse image-level attributes rather than segmenting the region of an attribute; or they do not address representing multiple attributes simultaneously on a single patch (important for e.g.~a detailed clothing representation including, e.g., category + texture + colour). 

Most other semantic attribute detection studies take a discriminative approach to learning for maximum detection accuracy \cite{vaquero2009attrib_surveil,layne2013attribReIdBook,satta2012appearanceSearch,Yamaguchi2012,Huizhong2012,Liu_MM_2014,kumar2008faceTracer}. In contrast, our generative approach provides advantages in more naturally supporting weakly-supervised learning, and domain transfer through Bayesian priors. Related generative models include \cite{wang_2014_bmva} which used unsupervised topic models to estimate saliency for re-id, and \cite{fu2012attribsocial,feng2014learnDict} which addressed data driven attribute discovery and learning. In a more related work, \cite{shi2014wslAttrib} proposed a weakly supervised approach which models attribute and object associations using a generalised IBP model.
However  \cite{shi2014wslAttrib} (i) is designed for a single domain rather than  for cross-domain use,  and (ii) has no notion of spatial coherence which is important for correctly segmenting fine person attribute details -- while we integrate a Markov Random Field (MRF) into the latent space of the IBP model for more accurate person segmentation. 

\noindent \textbf{Contributions:} Overall, the contributions of our model are as follows: (i) We introduce a generative framework for person attribute learning that can be learned from strongly or weakly annotated data or a mix; (ii) We show how to perform domain adaptation from fashion to surveillance domain; (iii) We demonstrate that the resulting representation is useful for both unsupervised and supervised re-identification where it achieves state-of-the-art, and near state-of-the-art results respectively; and (iv) The same framework enables better description-based person search compared against the existing discriminative modelling approaches. 


\vspace{-0.1cm}

\section{Semantic Representation Learning}
Our model, termed MRF-IBP, works with super-pixel segmented person images. Each super-pixel is associated with a $K$-dimensional latent binary vector whose elements (factors) indicate what set of attribute properties are possessed by that patch, and which of a potentially infinite set of background clutter types are present. The first $K_{s}$ factors are associated with known annotations, while the subsequent entries are free to model unannotated aspects of the images. Our pipeline exploits two datasets: an annotated auxiliary dataset, and an unannotated target dataset: \\
\textbf{Auxiliary Training:} First we train on the auxiliary/source dataset using weak (image-level) or strong (patch-level) supervision. The supervision is a binary vector  describing which attributes  appear in the image/patch. For strong supervision annotations $L^{(i)}_j$ are given for the first $K_{s}$ factors of patches $j$ in image $i$, and the model learns from this each of the $K_{s}$ factors' appearance. The $K_s$ supervised factors also include foreground versus background patch annotation. 
For weak supervision, annotations $L^{(i)}$ are given for the first $K_{s}$ factors of each image, and the model solves the (more challenging task) of learning both factor appearance and infer which image-level factors occur in which patches. \\
\textbf{Target Adaptation:} We then use the learned parameters from the auxiliary set as a prior, and adapt it to the target dataset without any supervision. The representation learned here is then used for re-identification and person-search.


\subsection{Model Formulation}

\begin{figure}[t]
\centering
\includegraphics[width=0.4\linewidth]{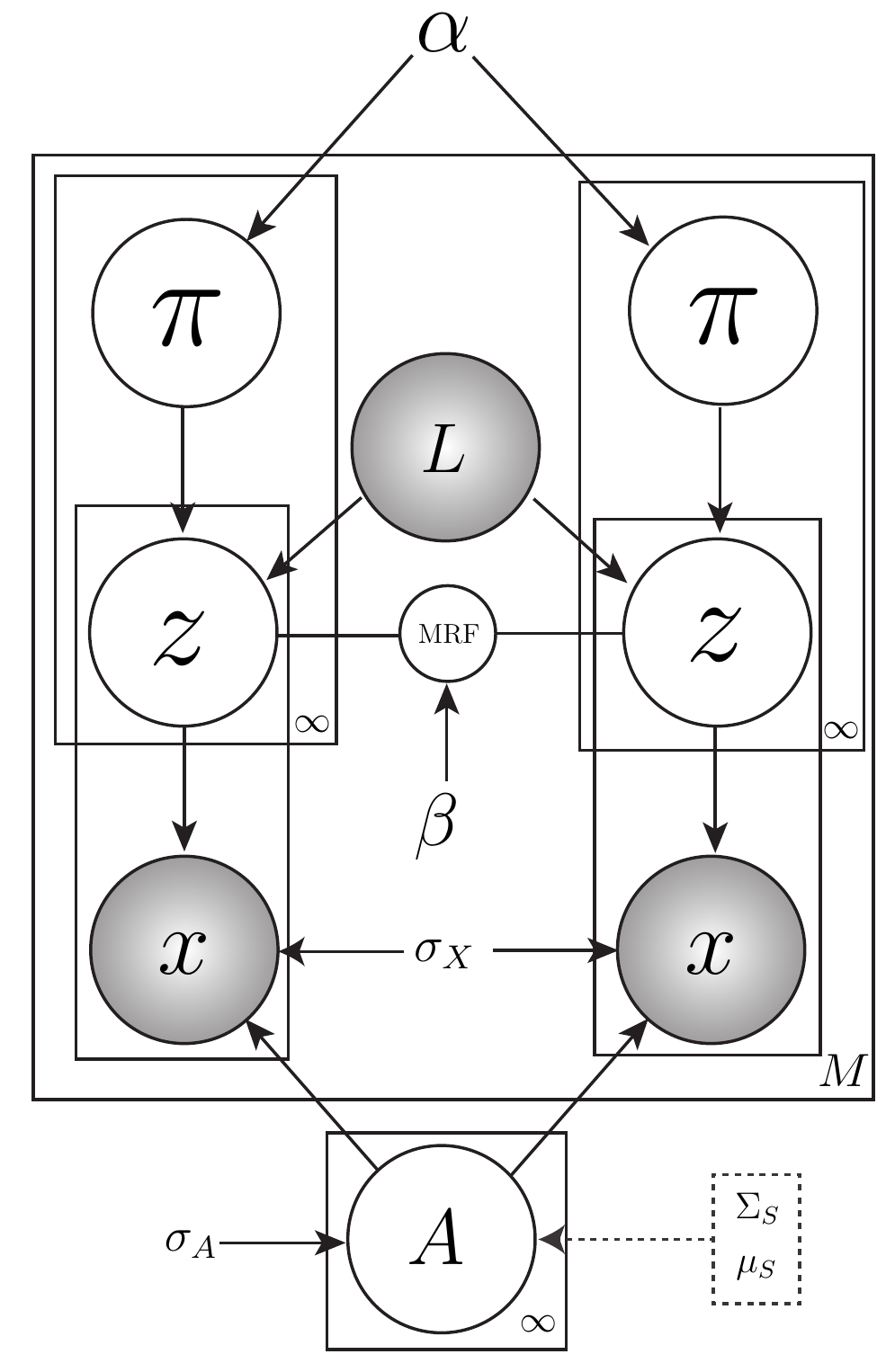}
\caption{The graphical model for MRF-IBP. Shaded nodes are observed. }
\label{fig:PGM}
\end{figure}

 Each image $i$ is represented as a bag of patches $\mathbf{X}^{(i)}=\{\mathbf{X}^{(i)}_{j\cdot}\}$, where $\bm{X}_{j\cdot}$ means the vector of row $j$ in matrix $\bm{X}$ and corresponds to a $D$-dimensional feature representation of each patch. Without supervision, the generative process for each image is as follows:\\

\noindent For each latent factor $k\in1\dots \infty$:

\begin{enumerate}

\item Draw an appearance distribution mean $\mathbf{A}_{k\cdot} \sim \mathcal{N}(\bm{\mu}_k,\Sigma_{k})$.

\end{enumerate}
For each image $i\in1\dots M$:
\begin{enumerate}

\item Draw the binary indicator matrix $\mathbf{Z}$ describing the factor activation for every patch:
\begin{eqnarray}
p(\mathbf{Z}^{(i)}|\alpha,\beta) & \propto & \frac{\alpha^{K_{+}}}{\prod_{j=1}^{N_i}K_{1}^{j}!}\cdot\exp-\alpha \sum^{N_i}_{j=1}\frac{1}{j}\nonumber\\
 &  & \hspace{-2cm}\cdot \prod_{k=1}^{K_{+}}\frac{\left(N_i-m^{(i)}_{k}\right)!(m^{(i)}_{k}-1)!}{N_i!}\nonumber\\
 &  & \hspace{-2cm}\cdot \prod_{k=1}^{K_{+}}\exp\left(\sum_{j=1}^{N_i}\beta\sum_{j'\in N(j)}I(z^{(i)}_{jk}=z^{(i)}_{j'k})\right)\label{eq:factorPrior}
\end{eqnarray}

 
\item For each super-pixel patch $j\in1\dots N_i$: Sample patch appearance: $\bm{X}_{j\cdot}^{(i)} \sim \mathcal{N}(\bm{Z}_{j\cdot}^{(i)}\bm{A},\sigma^2_X\bm{I}) $.

\end{enumerate}

\noindent Notations: $\mathcal{N}$ corresponds to Normal distribution with the specified parameters; $\alpha$ is the prior expected sparsity of annotations; $\beta$ is the coupling strength of the inter-patch MRF;  $\bm{\mu}_k$ and $\Sigma_A$ are the prior mean and covariance of each factor. $p(\bm{Z}^{(i)})$ in Eq.~(\ref{eq:factorPrior}) corresponds to our MRF-IBP prior. It expresses the IBP sampling of an unbounded number of factors in each patch (first two lines), which are  spatially correlated across patches by a Potts model MRF (last line) like \cite{Verbeek07regionclassification}. Here $K_+\geq K_s$ refers to the (inferred) number of active factors in the image, $K^j_1$ is the factor history \cite{Griffiths_2011}, and $m^{(i)}_k$ is the number of times each factor $k$ is active.

Denote the set of hidden variables by $\bm{H} = \{\bm{Z}^{(1)},\dots, \bm{Z}^{(M)}, \bm{A}\}$, observed images by $\bm{X} = \{\bm{X}^{(1)},\dots,\bm{X}^{(M)}\}$, and model parameters by $\bm{\Theta}=\{\alpha, \beta,\sigma_X, \Sigma_k, \bm{\mu}_k \}$. Then the joint  probability of the variables and data given the parameters is:
\begin{eqnarray}
p(\bm{H},\bm{X}|\Theta) & = & \prod_{k=1}^{\infty}p(\mathbf{A}_{k\cdot}|\bm{\mathbf{\mu}}_{k},\Sigma_{k})\nonumber\\
 &  & \hspace{-1.5cm}\cdot\prod_{i=1}^{M}p(\mathbf{Z}^{(i)};\alpha,\beta)\prod_{j=1}^{N_{i}}p(\mathbf{X}_{j\cdot}^{(i)}|\mathbf{Z}_{j\cdot}^{(i)},\mathbf{A},\sigma_{X})
\end{eqnarray}

Learning in our model aims to compute the posterior $ p(\bm{H} | \bm{X}, \bm{\Theta})$ for: discovering which factors (object/attributes) are active on each patch (inferring $\bm{Z}^{(i)}$), and learning the appearance of each factor (inferring $\mathbf{A}_{k\cdot}$).


\subsection{Model learning from the  Auxiliary Set}

To learn our MRF-IBP model, we exploit Gibbs sampling for approximate inference inspired by \cite{mrf_zhou_cvpr}. 
 For Gibbs sampling, we need to derive an update for each hidden variable conditional on the observations and all the other hidden variables. 

\noindent\textbf{Unsupervised Factor Updates:}\quad For all initialised factors $k$, we can sample the state of each latent factor $z_{jk}^{(i)}$ via:
\begin{align}
p(z_{jk}^{(i)}&=1|\bm{Z}_{-jk}^{(i)},\bm{X}_{j\cdot}^{(i)})\propto p(\bm{X}_{j\cdot}^{(i)}|\bm{Z}^{(i)})P(z_{jk}^{(i)}=1|\bm{Z}_{-jk}^{(i)}) \nonumber \\
&\hspace{-0.8cm}=  \frac{m^{(i)}_k-z^{(i)}_{jk}}{N_i}\cdot \exp \sum_{j'\in N(j)}\beta I(z^{(i)}_{jk}=z^{(i)}_{j'k})\nonumber\\
&\hspace{-0.8cm}\cdot\exp(-\frac{1}{2\sigma_X^ {2}}\mathrm{tr}(\bm{X}_{j.}^{(i)}-\bm{Z}_{j.}^{(i)}\bm{A})^T(\bm{X}_{j.}^{(i)}-\bm{Z}_{j.}^{(i)}\bm{A})) \label{eq:gibbsUpdate}
\end{align}
\noindent where $\bm{Z}_{-jk}^{(i)}$ denotes the entries of $\bm{Z}^{(i)}$ other than $\bm{Z}_{jk}^{(i)}$. To sample new latent factors, Poisson($\frac{\alpha}{N_i}$) is used as the expected number of new classes \cite{Griffiths_2011}.

\noindent\textbf{Supervised Factor Updates:}\quad Eq.~(\ref{eq:gibbsUpdate}) describes inference in the case where no supervision is available. If strong supervision $L^{(i)}_{jk}$ is available, Eq.~(\ref{eq:gibbsUpdate}) is replaced with $z^{(i)}_{jk} = L^{(i)}_{jk}$. If weak supervision $L^{(i)}_k$ is available Eq.~(\ref{eq:gibbsUpdate}) is replaced with $p(z_{jk}^{(i)}=1|\bm{Z}_{-jk}^{(i)},\bm{X}_{j\cdot}^{(i)})\propto p(\bm{X}_{j\cdot}^{(i)}|\bm{Z}^{(i)})P(z_{jk}^{(i)}=1|\bm{Z}_{-jk}^{(i)})\cdot L^{(i)}_{k}$ 

\noindent \textbf{Appearance Updates: }\quad In order to sample factor appearance $\bm{A}$, we compute its Gaussian posterior $p(\bm{A}|\bm{X},\bm{Z})$: 

\begin{eqnarray}
\mu_S &=&(\tilde{\bm{Z}}^T\tilde{\bm{Z}}+\frac{\sigma_X^2}{\sigma_A^2}I)^{-1}\tilde{\bm{Z}}^T\tilde{\bm{X}}\nonumber\\
\Sigma_S &=&\sigma_X^2(\tilde{\bm{Z}}^T\tilde{\bm{Z}}+\frac{\sigma_X^2}{\sigma_A^2}I)^{-1}\label{eq:auxAppear}
\end{eqnarray}

\noindent where $\tilde{\bm{Z}}$ and $\tilde{\bm{X}}$ are the matrices that vertically concatenate the factor matrix and patch feature matrix for all images. Here $\mu_S$ is the $K_+\times D$ matrix of appearance for each factor, and $\Sigma_S$ is the $K_+\times K_+$ matrix of variance parameters for each factor appearance. Since this is the auxiliary set, we have assumed an uninformative prior, i.e., $\bm{A}_k\sim\mathcal{N}(0,\sigma_A)$.

\begin{table*}[t]
\setlength{\tabcolsep}{0.2em}
\setlength\extrarowheight{1pt}
\centering
\footnotesize
\begin{tabular}{M{1cm} l || l  l  l  l ||  l l l l  ||  l l l l }
\hline
\multicolumn{2}{c||}{\multirow{2}{*}{Method}} & \multicolumn{4}{c||}{VIPeR}  &\multicolumn{4}{c||}{CUHK01} & \multicolumn{4}{c}{PRID450S} \\
\cline{3-14}
& & r=1 & r=5 & r=10 & r=20 & r=1 & r=5 & r=10 & r=20 & r=1 & r=5 & r=10 & r=20  \\
\hline
\hline

\parbox{5mm}{\multirow{4}{*}[-3pt]{\rotatebox{90}{single}}} &  \multicolumn{1}{|l||}{SDC \cite{zhao2013unsupervised}}  & 25.1 & 44.9 & 56.3 & 70.9 & 15.1  & 25.4  & 31.8 & 40.9 & 23.7 & 38.4 & 46.1 & 58.5\\
\cline{2-14}
& \multicolumn{1}{|l||}{GTS \cite{wang_2014_bmva}}  & 25.2 & 50.0 & 62.5 & 75.6 & - & - & - & - & -& -& -& -  \\
\cline{2-14}
& \multicolumn{1}{|l||}{SDALF \cite{Farenzena_2010CVPR}}  & 19.9 & 38.9 & 49.4 & 65.7 & 9.9 & 22.6 & 30.3 & 41.0   &17.4 & 30.9 & 40.8 & 55.2 \\
\cline{2-14}
& \multicolumn{1}{|l||}{Our unsupervised}  &  \textbf{27.7} & \textbf{55.3} & \textbf{68.3} & \textbf{79.7} & \textbf{23.3} & \textbf{35.8} & \textbf{46.6} & \textbf{60.7} &  \textbf{28.5} & \textbf{48.9}& \textbf{59.6} & \textbf{71.3} \\
\hhline{==============}
\parbox{5mm}{\multirow{2}{*}[-3pt]{\rotatebox{90}{fused}}} &  \multicolumn{1}{|l||}{SDC\underline{\ \ }{\it{Final}} (eSDC) \cite{zhao2013unsupervised}}  & 26.7 & 50.7 & 62.4 & 76.4 & 19.7 & 32.7  & 40.3 & 50.6 & 25.5 & 40.6 & 48.4 & 61.4  \\
\cline{2-14}
& \multicolumn{1}{|l||}{Our unsupervised\underline{\ \ }\it{Final}}  &  \textbf{29.3} & \textbf{52.7} & \textbf{66.8} & \textbf{79.7} & \textbf{22.4} & \textbf{35.9} & \textbf{47.9} & \textbf{64.5} &  \textbf{29.0} & \textbf{49.4} & \textbf{58.4} & \textbf{69.8}\\

\hline
\hline
\end{tabular}

\caption{Matching accuracy @ rank r (\%): unsupervised learning approaches. `-' indicates no result was reported and no code is available for implementation. The best results for single-cue and fused-cue methods are highlighted in bold separately.  }
\label{tab:unsupervised re-id}
\end{table*}

\subsection{Model Adaptation to the Target Set}

In the target set, there is no supervision, so Eq.~(\ref{eq:gibbsUpdate}) is used to update the latent factors. The appearance updates however are changed to reflect the top-down influence from the  learned auxiliary domain. Thus the target appearance $\mu_T$ is updated using the sufficient statistics from the source $\Sigma_S$ and $\mu_S$ (Eq.~(\ref{eq:auxAppear})) as the prior:

\begin{eqnarray}
\mu_T &=& \Sigma_T(\sigma_X^{-2}\tilde{\bm{Z}}^T\tilde{\bm{X}}+\Sigma_S^{-1}\mu_S)\nonumber\\
\Sigma_T &=&\sigma_X^2(\tilde{\bm{Z}}^T\tilde{\bm{Z}}+\sigma_X^2 \Sigma^{-1}_S)^{-1}\label{eq:targApp}
\end{eqnarray}




\section{Semantic Representation Applications}
After the semantic representation learning described previously, each target  image $i$ is now described by a binary factor matrix $\bm{Z}^{i}$ containing the inferred factor vector $\{\mathbf{z}_{j}\}_{j=1}^{N_i}$ for each super pixel $j$. We could use this representation directly, but find it convenient to convert it into a fixed-size representation per image. We thus generate multiple heat-maps $M_k$ per image representing the $k$th factor activation. Similar to \cite{xiong_eccv_2014}, we divide each image into 14 overlapping patches sampled on a 2$\times$7 regular grid with a 32$\times$32 window.\footnote{Note that the overlapping area between two neighbouring patches depends on the size of the image.} Each grid-patch is now represented by a $K$-dimensional attribute vector obtained by summing $\bm{z}_j$ for very pixel.

\subsection{Person Re-identification}


The proposed semantic person representation can be used for both unsupervised or supervised re-identification, according to whether a matching model is learned from the identity annotation of a given person re-id dataset.

\noindent\textbf{Unsupervised Matching:}\quad Each image is represented by 14 patches each with a $K$ dimensional descriptor. The person match is now converted to a semantic patch matching problem. We adopt a patch matching algorithm TreeCANN \cite{olonetsky2012treecann} to efficiently compute the distance between images.

\noindent\textbf{Supervised Matching:}\quad The 14 patch descriptors are concatenated to obtain an image-level descriptor. This is  used as input to a recent metric learning algorithm kLFDA \cite{xiong_eccv_2014}. 

\subsection{Person Search}
\label{sec:search}

Recall that K heat maps $M_k$ are generated from the K factors. The probability of factor $k$ appearing in an image  can be obtained by $max(M_k)$. When querying two or more factors, there are two possible query semantics: To query the probability of two factors both appearing anywhere in an image without preference for co-location (e.g., Coat + Bag), we use $max(M_k) \cdot max(M_{k'})$. In contrast, to query two factors that should \emph{simultaneously} appear in the same place (e.g., Blue-Jeans), we use $max(M_k \cdot M_{k'})$.

\begin{table*}[ht]
\setlength{\tabcolsep}{0.2em}
\setlength\extrarowheight{1pt}
\centering
\footnotesize
\begin{tabular}{M{1cm} l || l  l  l  l ||  l l l l  ||  l l l l }
\hline
\multicolumn{2}{c||}{\multirow{2}{*}{Method}} & \multicolumn{4}{c||}{VIPeR}  &\multicolumn{4}{c||}{CUHK01} & \multicolumn{4}{c}{PRID450S} \\
\cline{3-14}
& & r=1 & r=5 & r=10 & r=20 & r=1 & r=5 & r=10 & r=20 & r=1 & r=5 & r=10 & r=20  \\
\hline
\hline

\parbox{5mm}{\multirow{5}{*}[-5pt]{\rotatebox{90}{single}}} 
& \multicolumn{1}{|l||}{MLF \cite{zhao2014learning}}   & 29.1 & 52.3 & 65.9 & 79.9 & \textbf{34.3} & \textbf{55.1} & \textbf{65.0} & 74.9  & - & - & - & -\\
\cline{2-14}
& \multicolumn{1}{|l||}{KML \cite{xiong_eccv_2014}}  & 32.3 & 65.8 & 79.7 & 90.9 &  24.0 & 38.9 & 46.7 & 55.4 &32.4 & 54.4 & 62.4 & 69.6\\
\cline{2-14}
& \multicolumn{1}{|l||}{KISSME \cite{bischof2012largeScaleMetricLearn}}  & 19.6 & 48.0 & 62.2 & 77.0 &  8.4 & 25.1 & 38.7 & 50.2  & 26.5 & 47.8 & 57.6 & 68.5\\
\cline{2-14}
&  \multicolumn{1}{|l||}{SCNCD \cite{SalientColor_eccv_2014}} & \textbf{33.7} &  62.7 & 74.8 & 85.0 & - & - & - & -  & 41.5 & 66.6 & 75.9 & 84.4 \\ 
\cline{2-14}
& \multicolumn{1}{|l||}{FUSIA \cite{Ryan_bmvc_2014}} & 19.1 & 55.3& 73.5 & 84.8& 9.8 & 32.4 & 49.8 & 60.1  & - & - & - & - \\
\cline{2-14}
& \multicolumn{1}{|l||}{Our supervised}   &  31.1 & \textbf{68.6} &\textbf{82.8} & \textbf{94.9} & 32.7 & 51.2 & 64.4 & \textbf{76.3} & \textbf{43.1} & \textbf{70.5} & \textbf{78.2} & \textbf{86.3} \\
\hhline{==============}
\parbox{5mm}{\multirow{4}{*}[-5pt]{\rotatebox{90}{fused}}} & \multicolumn{1}{|l||}{KML\underline{\ \ }{\it{Final}} \cite{xiong_eccv_2014}}   & 36.1 & 68.7 & 80.1 & 85.6 & - & - & - & -  & - & - & - & - \\
\cline{2-14}
& \multicolumn{1}{|l||}{SCNCD\underline{\ \ }{\it{Final}} \cite{SalientColor_eccv_2014}}   & 37.8 & 68.6 & 81.0 & 90.5 & - & - & - & - & 41.6 & 68.9& \textbf{79.4} & \textbf{87.8} \\
\cline{2-14}
& \multicolumn{1}{|l||}{MLF\underline{\ \ }{\it{Final}} \cite{zhao2014learning}}   & \textbf{43.4} & \textbf{73.0} & 84.9 & 93.7 & - & - & - & -  & - & - & - & -\\
\cline{2-14}
& \multicolumn{1}{|l||}{Our supervised\underline{\ \ }\it{Final}}   & 41.6 & 71.9 & \textbf{86.2} & \textbf{95.1} &31.5 & 52.5 & \textbf{65.8} & \textbf{77.6} & \textbf{44.9} & \textbf{71.7} & 77.5 & 86.7\\

\hline
\hline
\end{tabular}

\caption{Matching accuracy @ rank r (\%): supervised learning approaches on re-id.}
\label{tab:supervised re-id}
\end{table*}

\section{Experiments}
\subsection{Datasets and Settings}
\begin{figure*}[t]
   \includegraphics[width=\linewidth]{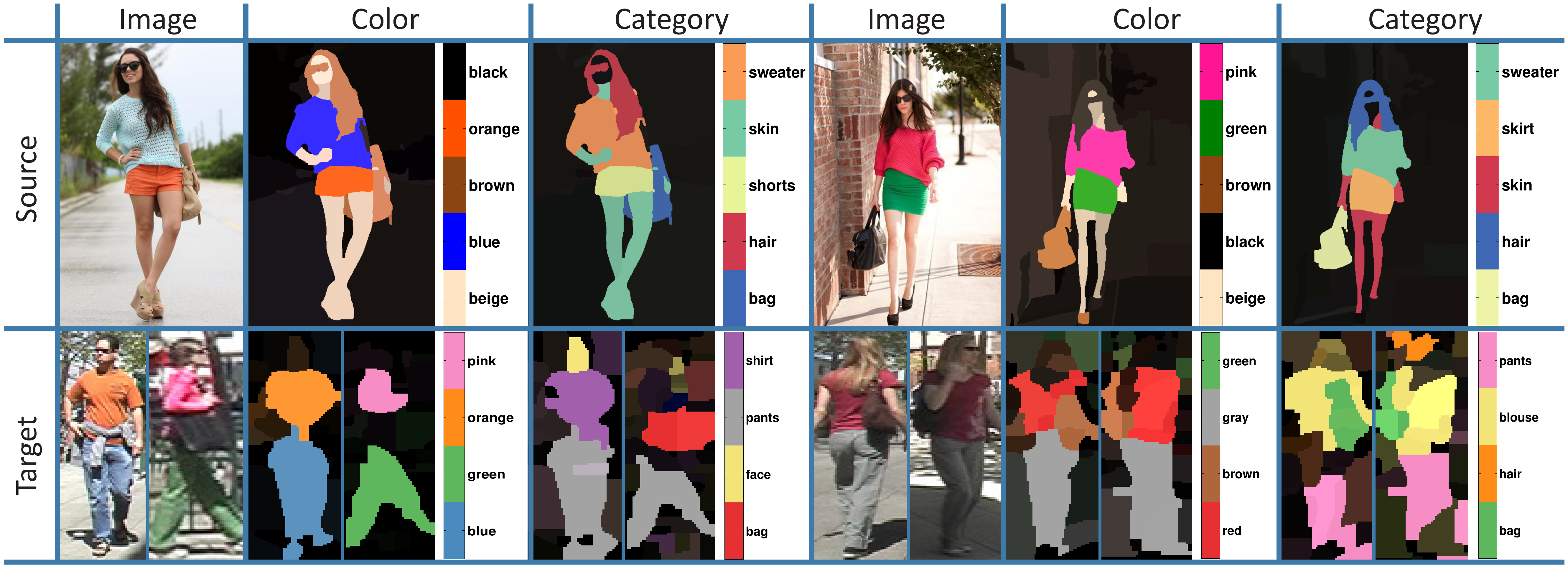}
   \caption{Visualisation of our model output. Each patch is colour-coded to show the inferred dominant attribute of two types.}
\label{fig:what is learned}
\end{figure*}

\noindent \textbf{Auxiliary Datasets:} Two  datasets are used as auxiliary sources. \textbf{Colourful-Fashion}  \cite{Liu_MM_2014} includes 2682 images. Pixel-level annotation is provided with  13 colour labels (e.g., brown, red) and 23 category labels (e.g., bag, T-shirt). Most of the images contain a single person with a relatively simple pose, against relatively clean background (see Fig.~\ref{fig:what is learned}). \textbf{Clothing-Attribute} \cite{Huizhong2012} includes 1,856 person images from social media sites, annotated with 26 attributes. Only image-level annotations are provided. However, it includes  6  texture attributes not included in Colourful-Fashion, so we include this auxiliary dataset mainly to enrich the representation with the 6 texture attributes.

\noindent \textbf{Target Datasets:} Four surveillance pedestrian datasets are used as target data. 
\textbf{VIPeR} \cite{viper07} contains two views of 632 pedestrians. 
All images are normalised to 128$\times$48. All images are also manually labeled by \cite{satta2012appearanceSearch} with 22 attributes, named VIPeR-Tag dataset \cite{satta2012appearanceSearch}. \textbf{CUHK01} \cite{li2012human} is captured with two camera views in a campus environment.
It contains 971 persons, with two images each. Images are normalised to 160 $\times$ 60. \textbf{PRID450S} \cite{prid_450s_2014}  is a recent and more realistic dataset, built on PRID 2011. It consists of 450 image pairs recorded from two static surveillance cameras. All images are normalised to 168 $\times$ 80.  \textbf{PETA} \cite{DENG_MM_2014} is a large-scale surveillance person attribute datatset that consists of 19000 images. Each image is labelled with 61 binary and 4 multi-class attributes, including colour, style etc. 

\noindent \textbf{Features:} 
We divide the image into super-pixels using a recent segmentation algorithm \cite{amfm_pami2011}. We represent each super-pixel as a vector using following features: (1)~Colour: We extract 3 dimensional colour descriptors from each pixel in both RGB and LAB colour space \cite{Gevers97colorbased,SalientColor_eccv_2014}. We run k-means to obtain 150 code words for each colour space. Pixels are quantised to the nearest centres in the visual vocabulary. The resulting descriptor for each super-pixel is the normalised histogram over visual words. (2)~SIFT: We compute 128 dimensional dense SIFT over a regular grid (4$\times$4 step size). Similar to Colour, we build a vocabulary of 300 words \cite{Liu_MM_2014}. A histogram is built from quantised local words within each super-pixel. (3)~Location: Following \cite{Liu_MM_2014,Yi_TPAMI_2013}, we consider a 2 dimensional coordinate of each super-pixel centroid as an absolute location feature. A relative location feature is defined by the distances between the centroid and each of 26 human key points generated by human pose estimation \cite{Yi_TPAMI_2013}, giving a 106 dimensional  location features. The final feature vector (706D) of each super-pixel is formed by concatenation of Colour (300D), SIFT (300D) and Location (106D). To compensate for the noise in the surveillance images, we also apply a rolling guidance filter \cite{rolling_filter_2014} before generating super-pixels. 

\noindent \textbf{Settings:}
For training the auxiliary datasets, we use $60$ supervised factors: 34 from Colourful-Fashion (12 colour + 22 category attributes), 6 (texture) from Clothing-Attribute, and 20 background factors (always off for foreground patches). Thus our model activates at least $K_+\geq K_s=60$ factors, although more may be used to explain un-annotated aspects of the data due to the use of IBP. We train by iterating Eqs.~(\ref{eq:gibbsUpdate}) and (\ref{eq:auxAppear}) for 2000 iterations. The supervision used varies across the strongly and weakly annotated auxiliary sets. Please see supplementary material for details. 
For transferring to the re-id datasets, we transfer the 60 auxiliary domain factors, and use $K_+\geq80$ by initialising a further 20 free factors randomly to accommodate new factors in the new domain. Any previously unseen unique aspects of the target domain can be modelled by these 20 factors.  We adapt the learned model to the target data by iterating Eqs.~(\ref{eq:gibbsUpdate}) and (\ref{eq:targApp}) for 100 iterations. We then take the first $K=80$ learned factors to produce an 80-dimensional patch representation (see Sec.~4) to be used in person re-id.  

\noindent \textbf{Baselines:}
In addition to comparing with start-of-the-art in person re-id and person search methods, we also consider alternative transfer  methods that could potentially generate an analogous representation to our framework: \textbf{SVM}/\textbf{MI-SVM}: SVM (as in \cite{layne2013attribReIdBook,satta2012appearanceSearch}) and Multi-Instance SVM \cite{Andrews03supportvector} are used to train patch-level attribute classifiers for strongly and weakly-labelled auxiliary data respectively. The learned SVMs can then be applied to estimate feature vectors for each target image patch similarly to our model. \textbf{DASA} \cite{fernando2013SA}: An unsupervised domain adaptation methods to address domain shift by aligning the source and target subspaces.

\noindent\textbf{Computational Cost:}
The complexity of our algorithm is $O(MN(K^2+KD))$ for $M$ images with $N$ super pixels, $K$ factors, and $D$-dimensional patch features. We run our algorithm on a PC with Intel 3.47 GHz CPU and 12GB RAM. In practice this corresponds to 1 to 2 minutes for 1000 images per iteration, depending on the number of super-pixels.

\subsection{Person Re-identification}

We first evaluate  person re-id performance against start-of-the-arts \cite{xiong_eccv_2014,VIPeR_2008,bischof2012largeScaleMetricLearn}.  We randomly divide the dataset into two equal, non-overlapping subsets for training and testing. We use the widely used accuracy at rank $k$ of Cumulative Match Characteristic (CMC) curves to quantify performance.  The results are obtained on VIPeR, CUHK01 and PRID450S datasets by averaging over 10 random splits. We distinguish using the suffix \_{\it{Final}} the common practice of use of an ensemble of methods or features with score or feature level fusion.


\noindent \textbf{Unsupervised Matching:} 
We compare our model to recent state-of-the-art approaches under an unsupervised setting (i.e.~no identity labels are used)  including SDALF \cite{Farenzena_2010CVPR}, eSDC~\cite{zhao2013unsupervised}, and GTS~\cite{wang_2014_bmva}. As shown in Table \ref{tab:unsupervised re-id}, our representation on its own significantly outperforms all other methods in all three datasets, and is not far off  the most competitive supervised methods (Table \ref{tab:supervised re-id}). When fused with SDALF as in \cite{zhao2013unsupervised}, performance improves further. See supplementary material for CMC curves and more comparisons.

\noindent \textbf{Supervised Matching:} 
Table \ref{tab:supervised re-id} compares our method in a supervised matching context against recent state-of-the-art including: MLF~\cite{zhao2014learning}, KML~\cite{xiong_eccv_2014}, KISSME~\cite{bischof2012largeScaleMetricLearn}, SCNCD~\cite{SalientColor_eccv_2014}, FUSIA~\cite{Ryan_bmvc_2014}. It shows that our approach achieves comparable or better performance to state-of-the-art, especially at higher rank (i.e. r=5,10,20). In this setting our final result is obtained by fusing with kLFDA \cite{xiong_eccv_2014}.

\noindent\textbf{Auxiliary Data:}
Here we evaluate the effects of various auxiliary data sources and annotations. Our full framework is learned with fully-supervised (f-F) Colourful-Fashion  and weakly-supervised (w-C) Clothing-Attribute  datasets. Table \ref{tab:dataSourcesl} (on ViPeR) shows that: (i) the different annotations in the two auxiliary datasets are combined synergistically, and weakly-annotated data can be used effectively (f-F+w-C $>$ f-F); and (ii) while capable of exploiting strong supervision where available, our framework does not critically rely on it (w-F+w-C close to f-F+w-C; w-F close to f-F). 

\begin{table}[h]
\scriptsize
\setlength{\tabcolsep}{0.2em}
\centering
\begin{tabular}{l|| l  l  l  l || l l l l }
\hline
\multirow{2}{*}{Auxiliary Data} &  \multicolumn{4}{c||}{Unsupervised} &  \multicolumn{4}{c}{Supervised} \\
\cline{2-9}
& r=1 & r=5 & r=10 & r=20 & r=1 & r=5 & r=10 & r=20 \\
\hline
\hline
w-F  &  18.3 & 38.3 & 49.5 & 62.9 & 26.2 & 58.2 & 71.1 & 83.4\\
\hline
f-F  & 25.4 &  51.4 &  63.9 & 75.3 & 29.4 & 64.9 & 78.8 & 91.7\\
\hline
w-F + w-C & 22.4 & 43.6 & 57.1 & 67.3  &  28.3 &62.2 & 75.8& 88.5\\
\hline
f-F + w-C  & \textbf{27.7} & \textbf{55.3}&\textbf{68.3} & \textbf{79.7} & \textbf{31.1} &  \textbf{68.6} & \textbf{82.8}&  \textbf{94.9} \\
\hline
\end{tabular}
\caption{Effects of auxiliary data source and annotation.}
\label{tab:dataSourcesl}
\end{table}

\vspace{0.2cm}
\noindent \textbf{Contributions of  Components:} To evaluate the contributions of each component of our framework, Table \ref{tab:abalation} summarises our  model performance  on ViPeR in 4 conditions: (1) Without MRF (NoMRF); (2) Direct transfer without adaptation (Eq.~\ref{eq:targApp}) (NoAdapt); (3) (1) $\&$ (2);  (4) Solely unsupervised target domain learning (NoTransfer). The results  show that each component (MRF modelling, transfer and adaptation)  contributes to the final performance.

\begin{table}[h]
\scriptsize
\setlength{\tabcolsep}{0.18em}
\centering
\begin{tabular}{l|| l  l  l  l || l l l l }
\hline
\multicolumn{1}{c||}{\multirow{2}{*}{Method}} &  \multicolumn{4}{c||}{Unsupervised} &  \multicolumn{4}{c}{Supervised} \\
\cline{2-9}
& r=1 & r=5 & r=10 & r=20 & r=1 & r=5 & r=10 & r=20 \\
\hline
\hline
NoMRF  &  23.3 & 47.4  & 59.1 &  70.9 &  28.0 & 62.1 & 75.2 & 87.2\\
\hline
NoAdapt  & 19.2 & 39.6  &  50.2 &  61.9 & 21.8 & 49.2 & 60.9 & 73.8 \\
\hline
NoMrfAdapt  & 17.7 & 36.2  &  45.5 & 54.8 & 20.2 & 46.0 &57.6 &70.9\\
\hline
NoTransfer & 9.5 & 20.5  & 26.9 & 35.6 & 14.3 & 32.9 & 41.7 & 52.5\\
\hline
Ours  & \textbf{27.7} & \textbf{55.3} & \textbf{68.3} & \textbf{79.7} & \textbf{31.1} &  \textbf{68.6} & \textbf{82.8} &  \textbf{94.9} \\
\hline
\end{tabular}
\caption{Contribution of each model component}
\label{tab:abalation}
\end{table}

\vspace{0.2cm}
\noindent \textbf{Alternative Transfer Approaches:}  Our model is compared against alternative SVM-based approaches. Table \ref{tab:SvmDaXfer} reveals that: (i) While (MI)SVMs can in principle deal with weakly or strongly supervised representation learning,  it  clearly under-performs our approach, and (ii) Although conventional feature-level domain adaptation (DASA \cite{fernando2013SA}) can improve the SVM performance, it is much less effective than our model-level adaptation.

\begin{table}[h]
\scriptsize
\setlength{\tabcolsep}{0.15em}
\centering
\begin{tabular}{ M{0.3cm} l|| l  l  l  l||  l l l l}
\hline
\multicolumn{2}{c||}{\multirow{2}{*}{Method}} &  \multicolumn{4}{c||}{Unsupervised} &  \multicolumn{4}{c}{Supervised} \\
\cline{3-10}
& & r=1 & r=5 & r=10 & r=20 & r=1 & r=5 & r=10 & r=20 \\
\hline
\hline
\parbox{5mm}{\multirow{3}{*}[-5pt]{\rotatebox{90}{w-F}}} & 
\multicolumn{1}{|l||}{MI-SVM} & 8.0 &  17.8  & 24.4 & 34.4 & 15.6 & 36.2 & 46.5 & 59.9\\
\cline{2-10}
& \multicolumn{1}{|l||}{DASA \cite{fernando2013SA}} & 12.2 & 25.8  & 33.9 & 43.7 &17.1 & 39.2 & 49.4 & 61.5\\
\cline{2-10}
&\multicolumn{1}{|l||}{Ours}   &  \textbf{18.3} & \textbf{38.3} & \textbf{49.5} & \textbf{62.9} & \textbf{26.2} & \textbf{58.2} & \textbf{71.1} & \textbf{83.4}\\
\hline
\hline
\parbox{5mm}{\multirow{3}{*}[-5pt]{\rotatebox{90}{f-F}}} & 
\multicolumn{1}{|l||}{SVM} & 13.2 &  29.6  & 40.3 & 55.4 & 17.4 & 40.5 & 51.9 & 66.8\\
\cline{2-10}
& \multicolumn{1}{|l||}{DASA \cite{fernando2013SA}} & 16.0 & 33.5  & 42.8 & 53.2 & 20.8 & 47.7 & 60.2 & 75.4\\
\cline{2-10}
&\multicolumn{1}{|l||}{Ours} & \textbf{25.4} &  \textbf{51.4} &  \textbf{63.9} & \textbf{75.3} & \textbf{29.4} & \textbf{64.9} & \textbf{78.8} & \textbf{91.7}\\
\hline

\end{tabular}
\caption{Comparing different transfer learning approaches}
\label{tab:SvmDaXfer}
\end{table}

\vspace{0.4cm}
\noindent\textbf{What is learned:}  Fig.~\ref{fig:what is learned} visualises after model learning how attributes are detected  given a new image. We visualise the top 5 most confident colour and category factors/attributes for each image in the test set of Colourful-Fashion. Our model can almost perfectly recognise and localise the attributes (top row). As expected, the inferred attributes are much more noisy for the re-id data (bottom row). However, overall they are accurate (e.g.~bags of different colours are detected), and crucially provide a much stronger representation than the even noisier low-level features.

\begin{figure*}[t]
   \includegraphics[width=\linewidth]{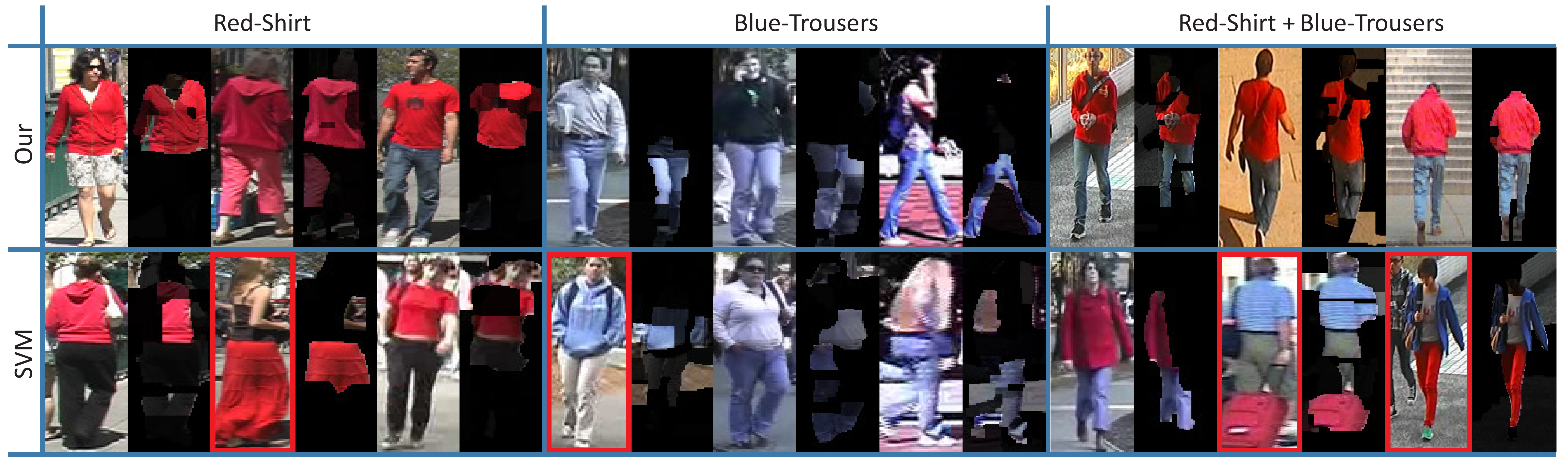}
   \caption{Person search qualitative results. The top ranked images for each query are shown. Red boxes are false detections.}
\label{fig:person search}
\end{figure*}

\subsection{Person Search} 
Although attribute-based query is a widely studied problem, there are few studies on person search \cite{Feris_icmr_2014,vaquero2009attrib_surveil} in surveillance. To evaluate description-based surveillance person search, we conduct experiments on VIPeR-Tag \cite{satta2012appearanceSearch} and PETA \cite{DENG_MM_2014}.  For both datasets, following \cite{satta2012appearanceSearch}, we randomly chose 50\% of the data for training (not used in our transfer framework, but used in other baselines) and the remaining for testing, and repeat this procedure 10 times to obtain average results.   Person Search is a retrieval task, so we evaluate the performance of each query with a precision-recall curve like \cite{satta2012appearanceSearch,Sivic03,ECCV_artem_2014}.

 \begin{figure}[h]
 
 \subfigure[2 terms queries, VIPeR]{
  \centering
  \includegraphics[height=0.4\linewidth]{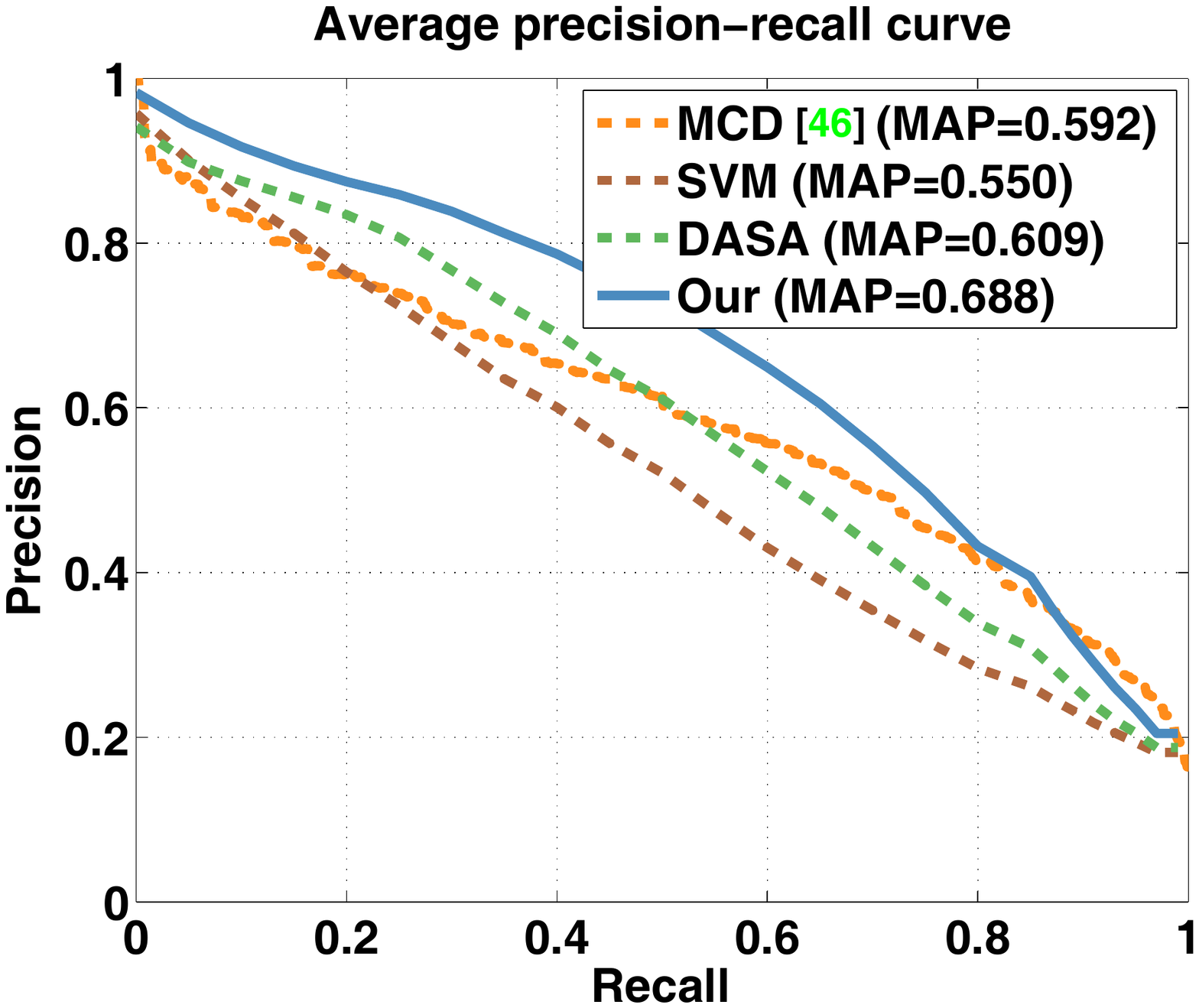} 
  \label{fig:sub1}
}%
\subfigure[4 terms queries, PETA]{
  \centering
   \includegraphics[height=0.4\linewidth]{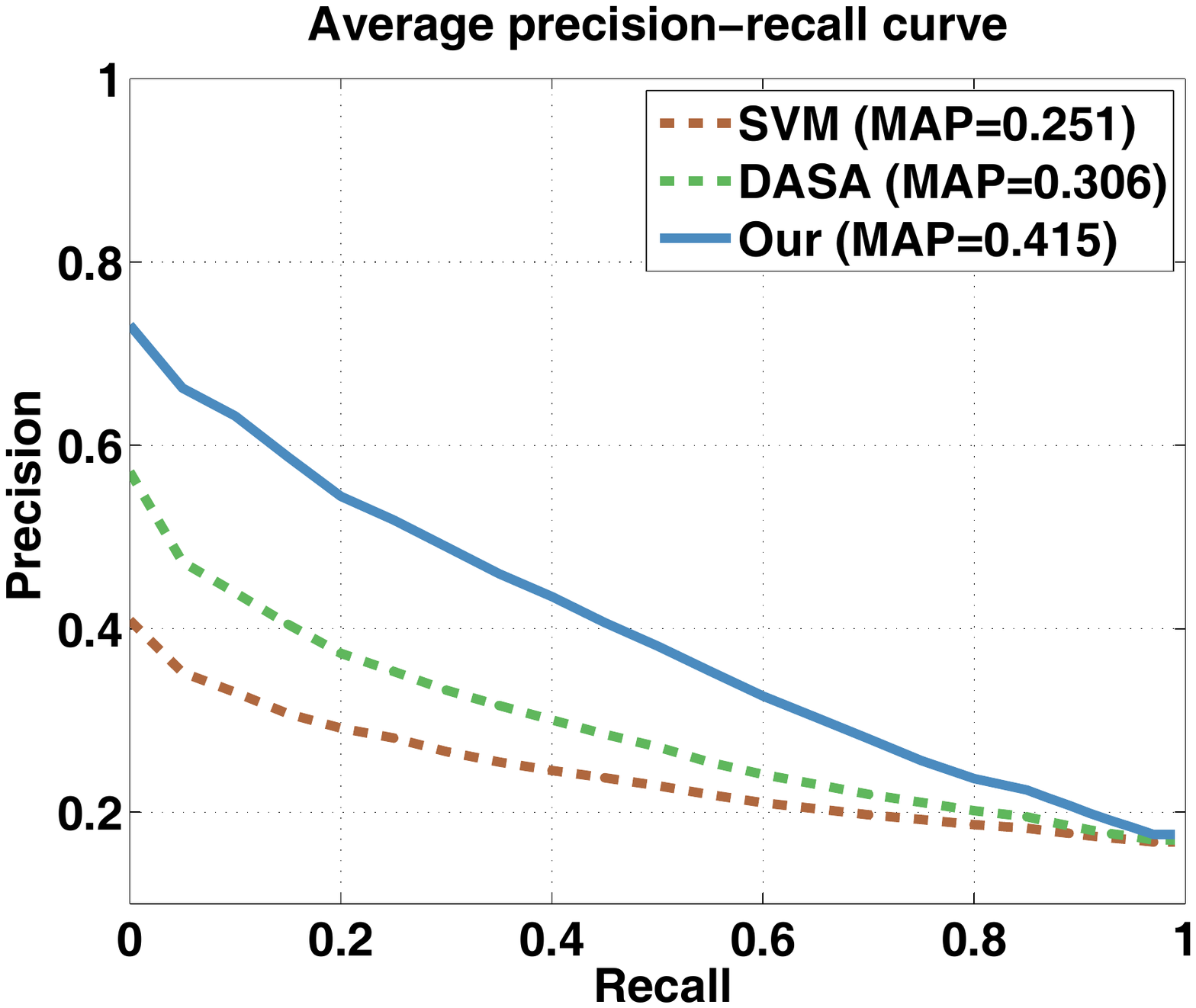}
  \label{fig:sub2}
}%
   \caption{Person search: comparison with state-of-the-art. }
\label{fig:searchSOA}
\end{figure}

 In VIPeR-Tag,  all 15 queries used in \cite{satta2012appearanceSearch} are contained in our source data attribute list. We can thus directly compare with the results in \cite{satta2012appearanceSearch}. The 15 queries are composed of a combination of an adjective (A) and a noun (N) (e.g.~Red-Shirt). To ensure two query terms co-exist in the same patch, we use $max(M_k \cdot M_{k'})$ to compute the  score (see Sec.~\ref{sec:search}). Fig.~\ref{fig:sub1} shows the average PR curves over all annotated queries, and it is clear that our method outperforms MCD \cite{satta2012appearanceSearch} \footnote{Various MCD versions are evaluated in \cite{satta2012appearanceSearch}. We compare with MCD$_1$ which gives the best MAP.}, SVM  \cite{layne2013attribReIdBook} and an unsupervised domain adaptation-based method DASA \cite{fernando2013SA}, even though no annotated VIPeR-Tag data is used for learning our model. Similar to \cite{Rastegari_CVPR13}, SVM scores have been calibrated by \cite{multiattrs_cvpr2012} before being fused to generate probability estimates of queries.  More detailed results (including  query-specific PR curves) are available in supplementary material.

In PETA, we consider a more challenging search task. Each query contains 4 terms of the form A-N+A-N (e.g.~Red-Shirt+Blue-Trousers).  We select all multi-class attribute labels  of PETA \cite{DENG_MM_2014}, including 11 colour (A) and 4 categories (N). In total 44 A-N combinations are generated and any two of them can form a 4-term query. Like \cite{Rastegari_CVPR13}, we randomly generate 200 4-term queries to evaluate the methods. Note that as the query form is A-N+A-N, the two query strategies in Sec.~\ref{sec:search} need to be combined to compute a score. Fig.~\ref{fig:sub2} shows that our method outperforms alternatives by a larger margin in this more difficult query setting.

Our model has two important advantages over the compared existing methods: (1) In order to better detect conjunctive person attributes such as ``Red-Shirt'', many existing methods  \cite{satta2012appearanceSearch} train a single attribute classifier for each combination of interest. This is not scalable because there will always be rare combinations that have too few instances to train a reliable classifier for; or at test time a combination may be required that no classifier has been trained for. By representing person attributes factorially, our model has no problem searching for combinations of attributes unseen at train time. (2) Because attributes are represented conjunctively at the patch-level, we can make complex queries such as (Black-Jeans + Blue-Shirt). An existing method such as the SVM-based one in \cite{layne2013attribReIdBook}, which uses image-level predictions for each attribute independently, may be confused by ``Blue-Jeans + Black-Shirt'' as an alternative. This explains the larger performance margin on PETA. Fig.~\ref{fig:person search} gives some qualitative illustration of these advantages. For example, Fig.~\ref{fig:person search} shows that the SVM-based model in \cite{layne2013attribReIdBook}, learned on each attribute separately at the image level, wrongly detects a  person with blue top when we query blue trousers. This limitation is more apparent for the more challenging ``Red-Shirt+Blue-Trousers'' query. In contrast, with patch-based joint attribute modelling, our model achieves much better results.

\vspace{10pt}
\section{Conclusion}

We have introduced a framework to generate semantic attribute representations of surveillance person images for re-id and search. Our framework exploits weakly and/or strongly annotated source data from other domains and transfers it with adaptation to obtain a good representation without any target domain annotation. The resulting patch-level semantic representation obtains competitive performance for supervised re-id, and state-of-the-art performance for unsupervised re-id -- which is the more practically relevant problem contexts since camera specific identity annotation is not scalable.  Moreover as a semantic representation it allows unification of re-id and person search within the same model.

\clearpage

{\small
\bibliographystyle{ieee}
\bibliography{egbib}

\begin{thebibliography}{10}\itemsep=-1pt

\bibitem{Andrews03supportvector}
S.~Andrews, I.~Tsochantaridis, and T.~Hofmann.
\newblock Support vector machines for multiple-instance learning.
\newblock In {\em NIPS}, 2003.

\bibitem{amfm_pami2011}
P.~Arbelaez, M.~Maire, C.~Fowlkes, and J.~Malik.
\newblock Contour detection and hierarchical image segmentation.
\newblock {\em TPAMI}, 2011.

\bibitem{ECCV_artem_2014}
A.~Babenko, A.~Slesarev, A.~Chigorin, and Lempitsky.
\newblock Neural codes for image retrieval.
\newblock In {\em ECCV}, 2014.

\bibitem{Cross_2012_wei}
W.~Bian, D.~Tao, and Y.~Rui.
\newblock Cross-domain human action recognition.
\newblock {\em Systems, Man, and Cybernetics, IEEE Transactions on}, 2012.

\bibitem{BourdevAttributesICCV11}
L.~Bourdev, S.~Maji, and J.~Malik.
\newblock Describing people: Poselet-based attribute classification.
\newblock In {\em ICCV}, 2011.

\bibitem{cao2010crossdata_action}
L.~Cao, Z.~Liu, and T.~S. Huang.
\newblock Cross-dataset action detection.
\newblock In {\em CVPR}, 2010.

\bibitem{Huizhong2012}
H.~Chen, A.~Gallagher, and B.~Girod.
\newblock Describing clothing by semantic attributes.
\newblock In {\em ECCV}, 2012.

\bibitem{abir_eccv_2014}
A.~Das, A.~Chakraborty, and A.~Roy-Chowdhury.
\newblock Consistent re-identification in a camera network.
\newblock In {\em ECCV}, 2014.

\bibitem{DENG_MM_2014}
Y.~Deng, P.~Luo, C.~C. Loy, and X.~Tang.
\newblock Pedestrian attribute recognition at far distance.
\newblock In {\em ACM Multimedia}, 2014.

\bibitem{Farenzena_2010CVPR}
M.~Farenzena, L.~Bazzani, A.~Perina, V.~Murino, and M.~Cristani.
\newblock Person re-identification by symmetry-driven accumulation of local
  features.
\newblock In {\em CVPR}, 2010.

\bibitem{feng2014learnDict}
J.~Feng, S.~Jegelka, S.~Yan, and T.~Darrell.
\newblock Learning scalable discriminative dictionary with sample relatedness.
\newblock In {\em CVPR}, 2014.

\bibitem{Feris_icmr_2014}
R.~Feris, R.~Bobbitt, L.~Brown, and S.~Pankanti.
\newblock Attribute-based people search: Lessons learnt from a practical
  surveillance system.
\newblock In {\em ICMR}, 2014.

\bibitem{fernando2013SA}
B.~Fernando, A.~Habrard, M.~Sebban, and T.~Tuytelaars.
\newblock Unsupervised visual domain adaptation using subspace alignment.
\newblock In {\em ICCV}, 2013.

\bibitem{Ferrari09}
V.~Ferrari, M.~Marin-Jimenez, and A.~Zisserman.
\newblock Pose search: {R}etrieving people using their pose.
\newblock In {\em CVPR}, 2009.

\bibitem{NIPS2007_3217}
V.~Ferrari and A.~Zisserman.
\newblock Learning visual attributes.
\newblock In {\em NIPS}, 2008.

\bibitem{fu2012attribsocial}
Y.~Fu, T.~Hospedales, T.~Xiang, and S.~Gong.
\newblock Attribute learning for understanding unstructured social activity.
\newblock In {\em ECCV}, 2012.

\bibitem{Gevers97colorbased}
T.~Gevers and A.~Smeulders.
\newblock Color based object recognition.
\newblock {\em Pattern Recognition}, 1997.

\bibitem{gong2012geodesicFlowDA}
B.~Gong, Y.~Shi, F.~Sha, and K.~Grauman.
\newblock Geodesic flow kernel for unsupervised domain adaptation.
\newblock In {\em CVPR}, 2012.

\bibitem{gong2014challenge}
S.~Gong, M.~Cristani, C.~C. Loy, and T.~M. Hospedales.
\newblock The re-identification challenge.
\newblock In S.~Gong, M.~Cristani, S.~Yan, and C.~C. Loy, editors, {\em Person
  Re-Identification}. Springer London, 2014.

\bibitem{gong2014reidBook}
S.~Gong, M.~Cristani, S.~Yan, and C.~C. Loy, editors.
\newblock {\em Person Re-Identification}.
\newblock Springer, 2014.

\bibitem{viper07}
D.~Gray, S.~Brennan, and H.~Tao.
\newblock Evaluating appearance models for recognition, reacquisition, and
  tracking.
\newblock In {\em IEEE International Workshop on Performance Evaluation for
  Tracking and Surveillance}, 2007.

\bibitem{VIPeR_2008}
D.~Gray and H.~Tao.
\newblock Viewpoint invariant pedestrian recognition with an ensemble of
  localized features.
\newblock In {\em ECCV}, 2008.

\bibitem{Griffiths_2011}
T.~L. Griffiths and Z.~Ghahramani.
\newblock The indian buffet process: An introduction and review.
\newblock {\em JMLR}, 2011.

\bibitem{khamis2014jointAttribute}
S.~Khamis, C.-H. Kuo, V.~K. Singh, V.~Shet, and L.~S. Davis.
\newblock Joint learning for attribute-consistent person re-identification.
\newblock In {\em ECCV Workshop on Visual Surveillance and Re-Identification},
  2014.

\bibitem{bischof2012largeScaleMetricLearn}
M.~Koestinger, M.~Hirzer, P.~Wohlhart, P.~M. Roth, and H.~Bischof.
\newblock Large scale metric learning from equivalence constraints.
\newblock In {\em CVPR}, 2012.

\bibitem{kumar2008faceTracer}
N.~Kumar, P.~Belhumeur, and S.~Nayar.
\newblock Facetracer: A search engine for large collections of images with
  faces.
\newblock In {\em ECCV}, 2008.

\bibitem{Kuo_2013}
C.-H. Kuo, S.~Khamis, and V.~Shet.
\newblock Person re-identification using semantic color names and rankboost.
\newblock In {\em WACV}, 2013.

\bibitem{layne2013attribReIdBook}
R.~Layne, T.~Hospedales, and S.~Gong.
\newblock {\em Person Re-identification}, chapter Attributes-based
  Re-identification.
\newblock Springer, 2014.

\bibitem{Ryan_bmvc_2014}
R.~Layne, T.~Hospedales, and S.~Gong.
\newblock Re-id: Hunting attributes in the wild.
\newblock In {\em BMVC}, 2014.

\bibitem{li2014clothesAttrib}
A.~Li, L.~Liu, K.~Wang, S.~Liu, and S.~Yan.
\newblock Clothing attributes assisted person re-identification.
\newblock {\em Circuits and Systems for Video Technology, IEEE Transactions
  on}, 2014.

\bibitem{li2012human}
W.~Li, R.~Zhao, and X.~Wang.
\newblock Human reidentification with transferred metric learning.
\newblock In {\em ACCV}, 2012.

\bibitem{li2014deepreid}
W.~Li, R.~Zhao, T.~Xiao, and X.~Wang.
\newblock Deepreid: Deep filter pairing neural network for person
  re-identification.
\newblock In {\em CVPR}, 2014.

\bibitem{ZhenliShiyu_CVPR2013}
Z.~Li, S.~Chang, F.~Liang, T.~S. Huang, L.~Cao, and J.~R. Smith.
\newblock Learning locally-adaptive decision functions for person verification.
\newblock In {\em CVPR}, June 2013.

\bibitem{Liu_MM_2014}
S.~Liu, J.~Feng, C.~Domokos, H.~Xu, J.~Huang, Z.~Hu, and S.~Yan.
\newblock Fashion parsing with weak color-category labels.
\newblock {\em Multimedia, IEEE Transactions on}, 2014.

\bibitem{Liu2012}
S.~Liu, Z.~Song, G.~Liu, C.~Xu, H.~Lu, and S.~Yan.
\newblock Street-to-shop: Cross-scenario clothing retrieval via parts alignment
  and auxiliary set.
\newblock In {\em CVPR}, 2012.

\bibitem{Liu_2014_CVPR}
X.~Liu, M.~Song, D.~Tao, X.~Zhou, C.~Chen, and J.~Bu.
\newblock Semi-supervised coupled dictionary learning for person
  re-identification.
\newblock In {\em CVPR}, 2014.

\bibitem{liu2012attribTopic}
X.~Liu, M.~Song, Q.~Zhao, D.~Tao, C.~Chen, and J.~Bu.
\newblock Attribute-restricted latent topic model for person re-identification.
\newblock {\em Pattern Recognition}, 2012.

\bibitem{Ma_2013_ICCV}
A.~J. Ma, P.~C. Yuen, and J.~Li.
\newblock Domain transfer support vector ranking for person re-identification
  without target camera label information.
\newblock In {\em ICCV}, 2013.

\bibitem{mccallum1998activeem}
A.~McCallum and K.~Nigam.
\newblock Employing em and pool-based active learning for text classification.
\newblock In {\em ICML}, 1998.

\bibitem{olonetsky2012treecann}
I.~Olonetsky and S.~Avidan.
\newblock Treecann - k-d tree coherence approximate nearest neighbor algorithm.
\newblock In {\em ECCV}, 2012.

\bibitem{Konstantina_tpami_2014}
K.~Palla, D.~Knowles, and Z.~Ghahramani.
\newblock Relational learning and network modelling using infinite latent
  attribute models.
\newblock {\em TPAMI}, 2014.

\bibitem{pan2009transfer_survey}
S.~J. Pan and Q.~Yang.
\newblock A survey on transfer learning.
\newblock {\em TKDE}, 2010.

\bibitem{Pedagadi_2013_CVPR}
S.~Pedagadi, J.~Orwell, S.~Velastin, and B.~Boghossian.
\newblock Local fisher discriminant analysis for pedestrian re-identification.
\newblock In {\em CVPR}, 2013.

\bibitem{Rastegari_CVPR13}
M.~Rastegari, A.~Diba, D.~Parikh, and A.~Farhadi.
\newblock Multi-attribute queries: To merge or not to merge?
\newblock In {\em CVPR}, 2013.

\bibitem{roth2014jointAttributeLearning}
J.~Roth and X.~Liu.
\newblock On the exploration of joint attribute learning for person
  re-identification.
\newblock In {\em ACCV}, 2014.

\bibitem{prid_450s_2014}
P.~M. Roth, M.~Hirzer, M.~Koestinger, C.~Beleznai, and H.~Bischof.
\newblock Mahalanobis distance learning for person re-identification.
\newblock In {\em Person Re-Identification}. Springer, 2014.

\bibitem{saenko2010domainAdapt}
K.~Saenko, B.~Kulis, M.~Fritz, and T.~Darrell.
\newblock Adapting visual category models to new domains.
\newblock In {\em ECCV}, 2010.

\bibitem{satta2012appearanceSearch}
R.~Satta, G.~Fumera, and F.~Roli.
\newblock People search with textual queries about clothing appearance
  attributes.
\newblock In {\em Person Re-Identification}. Springer, 2014.

\bibitem{multiattrs_cvpr2012}
W.~Scheirer, N.~Kumar, P.~N. Belhumeur, and T.~E. Boult.
\newblock Multi-attribute spaces: Calibration for attribute fusion and
  similarity search.
\newblock In {\em CVPR}, 2012.

\bibitem{shi2014wslAttrib}
Z.~Shi, Y.~Yang, T.~M. Hospedales, and T.~Xiang.
\newblock Weakly supervised learning of objects, attributes and their
  associations.
\newblock In {\em ECCV}, 2014.

\bibitem{Sivic03}
J.~Sivic and A.~Zisserman.
\newblock Video google: {A} text retrieval approach to object matching in
  videos.
\newblock In {\em ICCV}, 2003.

\bibitem{vaquero2009attrib_surveil}
D.~Vaquero, R.~Feris, D.~Tran, L.~Brown, A.~Hampapur, and M.~Turk.
\newblock Attribute-based people search in surveillance environments.
\newblock In {\em WACV}, 2009.

\bibitem{Verbeek07regionclassification}
J.~Verbeek and B.~Triggs.
\newblock Region classification with markov field aspect models.
\newblock In {\em CVPR}, 2007.

\bibitem{vezzani2013reidSurvey}
R.~Vezzani, D.~Baltieri, and R.~Cucchiara.
\newblock People re-identification in surveillance and forensics: a survey.
\newblock {\em ACM Computing Surveys}, Dec. 2013.

\bibitem{wang_2014_bmva}
H.~Wang, S.~Gong, and T.~Xiang.
\newblock Unsupervised learning of generative topic saliency for person
  re-identification.
\newblock In {\em BMVC}, 2014.

\bibitem{xiong_eccv_2014}
F.~Xiong, M.~Gou, O.~Camps, and M.~Sznaier.
\newblock Person re-identification using kernel-based metric learning methods.
\newblock In {\em ECCV}, 2014.

\bibitem{Xu_2013_ICCV}
Y.~Xu, L.~Lin, W.-S. Zheng, and X.~Liu.
\newblock Human re-identification by matching compositional template with
  cluster sampling.
\newblock In {\em ICCV}, 2013.

\bibitem{Yamaguchi_2013_ICCV}
K.~Yamaguchi, M.~Hadi~Kiapour, and T.~L. Berg.
\newblock Paper doll parsing: Retrieving similar styles to parse clothing
  items.
\newblock In {\em ICCV}, 2013.

\bibitem{Yamaguchi2012}
K.~Yamaguchi, M.~H. Kiapour, L.~E. Ortiz, and T.~L. Berg.
\newblock Parsing clothing in fashion photographs.
\newblock In {\em CVPR}, 2012.

\bibitem{Yang_2014_CVPR}
W.~Yang, P.~Luo, and L.~Lin.
\newblock Clothing co-parsing by joint image segmentation and labeling.
\newblock In {\em CVPR}, 2014.

\bibitem{Yi_TPAMI_2013}
Y.~Yang and D.~Ramanan.
\newblock Articulated human detection with flexible mixtures of parts.
\newblock {\em TPAMI}, 2013.

\bibitem{SalientColor_eccv_2014}
Y.~Yang, J.~Yang, J.~Yan, S.~Liao, D.~Yi, and S.~Li.
\newblock Salient color names for person re-identification.
\newblock In {\em ECCV}, 2014.

\bibitem{rolling_filter_2014}
Q.~Zhang, X.~Shen, L.~Xu, and J.~Jia.
\newblock Rolling guidance filter.
\newblock In {\em ECCV}, 2014.

\bibitem{zhao2013person}
R.~Zhao, W.~Ouyang, and X.~Wang.
\newblock Person re-identification by salience matching.
\newblock In {\em ICCV}, 2013.

\bibitem{zhao2013unsupervised}
R.~Zhao, W.~Ouyang, and X.~Wang.
\newblock Unsupervised salience learning for person re-identification.
\newblock In {\em CVPR}, 2013.

\bibitem{zhao2014learning}
R.~Zhao, W.~Ouyang, and X.~Wang.
\newblock Learning mid-level filters for person re-identfiation.
\newblock In {\em CVPR}, 2014.

\bibitem{cvpr_ZhengGX12}
W.-S. Zheng, S.~Gong, and T.~Xiang.
\newblock Transfer re-identification: From person to set-based verification.
\newblock In {\em CVPR}, 2012.

\bibitem{mrf_zhou_cvpr}
B.~Zhou, X.~Wang, and X.~Tang.
\newblock Random field topic model for semantic region analysis in crowded
  scenes from tracklets.
\newblock In {\em CVPR}, 2011.

\bibitem{Zitnick_2013_ICCV}
C.~L. Zitnick and D.~Parikh.
\newblock Bringing semantics into focus using visual abstraction.
\newblock In {\em CVPR}, 2013.

\end{thebibliography}
}

\end{document}